\documentclass[letterpaper]{article}

\usepackage[preprint]{aaai2027}
\usepackage[hyphens]{url}
\usepackage{graphicx}
\usepackage{natbib}
\usepackage{caption}
\usepackage{amsmath}
\usepackage{amssymb}
\usepackage{booktabs}

\graphicspath{{assets/}}

\newif\ifhumanstudyresults
\humanstudyresultstrue

\title{FeatFix: Reuse What You Verify through Local Exact-Feature Correction for Faster Cached Diffusion Inference}
\author{Hanshuai Cui\textsuperscript{2,1}, Zhiqing Tang\textsuperscript{1}, Zhi Yao\textsuperscript{2,1}, Qianli Ma\textsuperscript{1}, Fanshuai Meng\textsuperscript{1}, Weijia Jia\textsuperscript{1}}
\affiliations{\textsuperscript{1}Institute of Artificial Intelligence and Future Networks, Beijing Normal University, Zhuhai 519087, China\\
\textsuperscript{2}School of Artificial Intelligence, Beijing Normal University, Beijing 100875, China}

\begin{document}
\maketitle

\begin{abstract}
Diffusion models are widely used to generate high-quality images and videos, but their iterative denoising process remains computationally intensive. A growing class of training-free accelerators reduces this cost by reusing cached intermediate features or forecasting future ones. To control draft drift, these methods sometimes compute an exact block feature for verification. Yet the resulting exact feature is typically used only to measure discrepancy or guide a later decision and is then discarded. We find that this previously computed feature can instead be reused for correction. Forwarding it at the verification site resets the local draft residual and reduces downstream feature error. Based on this observation, we introduce FeatFix, a local exact-feature correction method for cached diffusion inference. FeatFix operates at a fixed sparse set of layer--timestep sites. At each selected site, it replaces the complete draft block output with the exact output computed from the same incoming state, avoiding token- or channel-level partial replacement and full-timestep recomputation. Experiments across four image and video backbones show that FeatFix consistently accelerates generation, achieving a speedup of up to $6.70\times$ over Vanilla while maintaining competitive output quality.
\end{abstract}

\section{Introduction}

Diffusion models have become a widely used paradigm for generating high-fidelity images and increasingly coherent videos~\cite{ho2020ddpm,dhariwal2021adm,saharia2022imagen,rombach2022ldm}. The same iterative generation principle extends across score-based formulations~\cite{song2021sde}, flow matching~\cite{lipman2023flowmatching}, and rectified flow~\cite{liu2023rectifiedflow}. Modern diffusion transformers (DiTs) improve scalability by replacing convolutional denoisers with deep stacks of attention and MLP blocks~\cite{peebles2023dit,esser2024sd3}. Their inference cost, however, remains tied to repeatedly evaluating those blocks over many denoising timesteps. As model width, depth, spatial resolution, and video length grow, this repeated computation makes latency and resource demands central obstacles to interactive image and video generation.

\begin{figure}[t]
\centering
\includegraphics[width=\columnwidth]{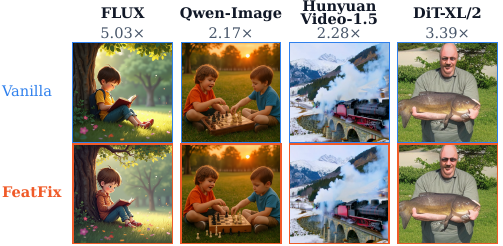}
\caption{Cross-backbone comparisons. Each column pairs Vanilla with FeatFix at the reported speedup across FLUX, Qwen-Image, HunyuanVideo-1.5, and DiT-XL/2.}
\label{fig:teaser}
\end{figure}

Existing acceleration strategies fall broadly into training-based and training-free categories. Training-based acceleration distills a long trajectory into fewer steps or optimizes a consistency objective~\cite{salimans2022progressive,song2023consistency}. It reduces network evaluations but requires additional data, optimization, and a new model artifact, often for each backbone or deployment setting. Training-free acceleration instead operates on a pretrained checkpoint without modifying or retraining its parameters. Numerical solvers shorten the trajectory~\cite{karras2022edm,lu2022dpmsolver}, while feature-level methods exploit temporal redundancy inside the denoiser. They reuse cached activations, forecast future features from their history, or selectively refresh exact computation~\cite{deepcache2024,fora2024,teacache2025,taylorseer2025,speca2025}. Compatibility with pretrained checkpoints lowers adaptation cost when retraining is unavailable. The underlying draft policy remains responsible for deciding where ordinary inference should reuse, forecast, or fully refresh a feature.

Reuse and forecasting replace exact intermediate states with cheaper draft features, so approximation error can accumulate across timesteps and transformer blocks. To monitor this drift, verification-based accelerators occasionally compute a draft block output and its exact counterpart from the \emph{same incoming state}~\cite{speca2025}. This check already incurs the transformer computation needed to produce a trustworthy feature. Nevertheless, the Verify control may log the discrepancy, discard the already-computed exact feature, and continue forwarding the draft. The check thus diagnoses the current error without correcting the current state. We find that forwarding the exact tensor at the verification site eliminates the local same-input residual by construction and reduces downstream feature error.

Turning an already-computed exact feature into a practical correction raises two key challenges. \emph{(C1) How can it correct the current draft without reducing the speedup?} Replacing only selected tokens or channels leaves part of the block error uncorrected, while recomputing the full timestep repeats many transformer blocks. The correction should therefore reuse the complete exact block output without any additional exact computation. \emph{(C2) How can a correction at only a few sites remain effective downstream?} Exact replacement removes the same-input residual at the corrected block, but later layers and timesteps may weaken this local gain because the trajectory has already diverged. We therefore apply corrections at fixed, auditable layer--timestep sites and evaluate their effects against a matched-compute Verify control.

We introduce FeatFix to address these challenges. It keeps the underlying draft policy unchanged and acts only at a fixed set of layer--timestep verification sites. At each site, FeatFix forwards the complete exact block output computed from the same incoming state instead of the draft. Because the matched Verify control already computes this exact output, FeatFix removes the local residual without additional block evaluation or full-timestep recomputation. We record full refreshes, paid checks, and forwarded corrections separately to distinguish correction from scheduling. Matched Verify controls isolate the effect of forwarding the exact feature, while downstream feature traces test whether that effect persists beyond the corrected site. We evaluate FeatFix on FLUX.1-dev, Qwen-Image, HunyuanVideo-1.5, and DiT-XL/2 across image and video generation. It reaches up to $6.70\times$ speedup over Vanilla while maintaining competitive output quality. In the controlled FLUX study, FeatFix improves ImageReward over matched Verify, passes the LPIPS non-inferiority guard, and operates at essentially the same latency as Verify. Figure~\ref{fig:teaser} shows examples. Our contributions are threefold.
\begin{itemize}
\item We formulate paid exact-feature reuse as a local correction principle and show why replacing the complete block output is the compute-consistent unit for removing the verified same-input residual.
\item We introduce FeatFix, a draft-policy-independent framework for fixed, sparse, and auditable correction with matched Verify controls and explicit full-refresh, paid-check, and forwarded-correction accounting.
\item We validate FeatFix across four image and video backbones, achieving up to $6.70\times$ speedup while preserving quality. Feature traces show downstream error reduction, and a controlled evaluation confirms quality gains at the same computational cost.
\end{itemize}

\begin{figure}[!t]
\centering
\begin{tabular}{@{}p{0.485\columnwidth}@{\hspace{0.03\columnwidth}}p{0.485\columnwidth}@{}}
\centering\includegraphics[width=\linewidth]{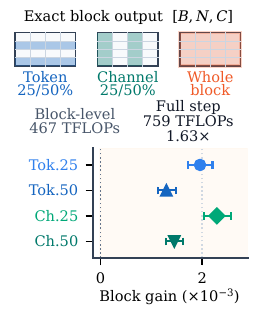} & \centering\includegraphics[width=\linewidth]{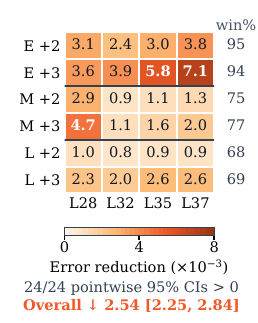}\tabularnewline[2pt]
\centering{\small\textbf{(a) Whole block is the efficient unit.}} & \centering{\small\textbf{(b) Correction persists downstream.}}\tabularnewline
\end{tabular}
\caption{Whole-block correction and downstream propagation. Whole-block forwarding reuses the paid exact tensor without partial residuals or whole-timestep recomputation, and its error reduction persists at later layers.}
\label{fig:motivation-correction}
\end{figure}

\begin{figure*}[!t]
\centering
\includegraphics[width=0.985\textwidth]{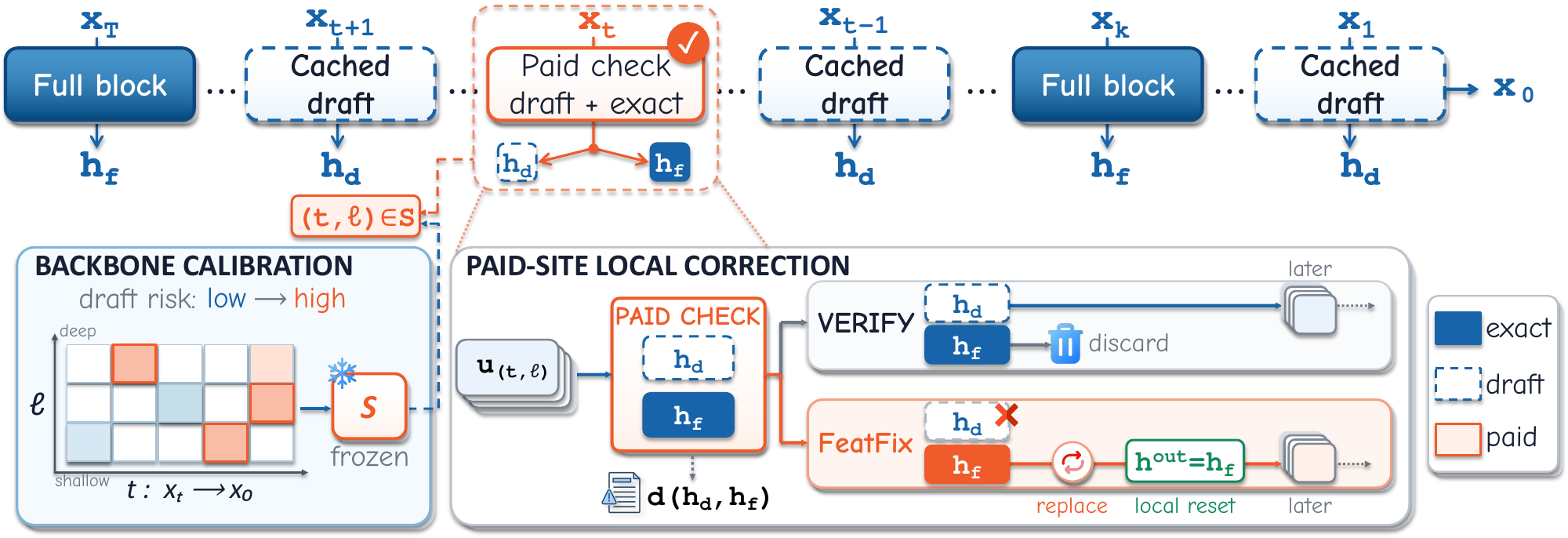}
\caption{FeatFix framework. At sites defined by a frozen layer--timestep schedule, FeatFix forwards the already-computed exact whole-block feature while the matched Verify arm discards it.}
\label{fig:method_overview}
\end{figure*}

\section{Related Work}

\paragraph{Feature reuse and adaptive caching.} Training-free accelerators exploit temporal redundancy inside a pretrained denoiser. DeepCache~\cite{deepcache2024} introduced intermediate-feature reuse without retraining. FORA~\cite{fora2024} and $\Delta$-DiT~\cite{deltadit2024} use fixed or structured refresh schedules, whereas ToCa~\cite{toca2025} selects reusable content at token granularity. TeaCache~\cite{teacache2025} and DiCache~\cite{dicache2025} use proxy errors or feature dynamics to trigger computation. HiCache~\cite{hicache2026}, ScalingCache~\cite{scalingcache2026}, and DBCache~\cite{dbcache2025} introduce hierarchical, scaled, or block-wise reuse. Video methods exploit magnitude, attention, or conditional redundancy~\cite{magcache2025,pab2025,fastercache2025}. These policies decide what to reuse and when to refresh; FeatFix preserves those decisions and acts only when a paid check has produced both draft and exact block outputs.

\paragraph{Forecasting, verification, and local correction.} TaylorSeer~\cite{taylorseer2025} forecasts DiT features from temporal derivatives, while SpeCa~\cite{speca2025} combines a Taylor draft with speculative checks and uses discrepancy to accept or reject predicted states and adjust later computation. FeatFix instead asks whether the exact feature produced by a check should replace the current draft or serve only as a discrepancy signal. It defines neither a forecasting rule nor an online acceptance policy and applies to both reuse and forecast drafts. Its matched Verify control uses the same draft, incoming state, exact block evaluation, and fixed sites but discards the exact output. Forwarding that output isolates local correction from scheduling and removes the selected block's same-input residual, although the remaining trajectory may still differ from Vanilla.

\section{Method}

\subsection{Preliminaries}

\noindent\textbf{Diffusion and DiT denoising.} Diffusion, flow-matching, and rectified-flow models denoise a sequence of noisy states~\cite{song2021sde,lipman2023flowmatching,liu2023rectifiedflow}. DiT implements the denoiser with $L$ transformer blocks~\cite{peebles2023dit}. For block $\ell$ at timestep $t$, let $u_{t,\ell}$ be its input and
\begin{equation}
h^{\mathrm f}_{t,\ell}=B_\ell(u_{t,\ell};t,c),
\label{eq:full-feature}
\end{equation}
its exact output, where $c$ denotes text or class conditioning. A generation trajectory therefore contains a two-dimensional grid of layer--timestep sites $(t,\ell)$ rather than only a sequence of denoising steps.

\medskip
\noindent\textbf{Feature reuse and forecasting.} Training-free accelerators avoid evaluating Eq.~\eqref{eq:full-feature} at every site. A reuse policy forwards a stored feature, e.g., $h^{\mathrm d}_{t,\ell}=h^{\mathrm f}_{s,\ell}$ for an earlier refresh $s$. TaylorSeer first showed that future DiT features can instead be forecast from their temporal history~\cite{taylorseer2025}. In generic form, an order-$K$ Taylor draft is
\begin{equation}
h^{\mathrm d}_{t,\ell}=\sum_{k=0}^{K}
\frac{(t-s)^k}{k!}\,\widehat{\Delta^{(k)}h}_{s,\ell},
\label{eq:taylor-draft}
\end{equation}
with finite differences estimated at exact refreshes. We denote any such base draft policy by $\mathcal D$. Importantly, $\mathcal D$ decides when ordinary sites reuse, predict, or refresh.

\medskip
\noindent\textbf{Paid verification and hidden drift.} At skipped sites, $h^{\mathrm f}_{t,\ell}$ is unavailable, so the local error remains hidden. A paid verification exposes it by evaluating the draft and exact block output from the same incoming state:
\begin{equation}
h^{\mathrm d}_{t,\ell}=\mathcal D(u_{t,\ell}),\qquad
h^{\mathrm f}_{t,\ell}=B_\ell(u_{t,\ell};t,c).
\label{eq:same-input-pair}
\end{equation}
We log their normalized relative-$L_1$ discrepancy
\begin{equation}
d_{\mathrm{rel}\text{-}L_1}(h_{\rm d},h_{\rm f})=
\frac{\left\lVert h_{\rm d}-h_{\rm f}\right\rVert_1}
{\left\lVert h_{\rm f}\right\rVert_1+\epsilon}.
\label{eq:relative-l1}
\end{equation}
Here $\epsilon>0$ ensures numerical stability. This audit signal does not route the current sample: Verify may use it for later decisions but still forwards $h^{\mathrm d}_{t,\ell}$ and discards the paid $h^{\mathrm f}_{t,\ell}$. FeatFix asks whether that exact result should instead repair the current state.

\begin{table*}[!t]
\centering
\small
\begin{tabular}{@{}l@{\hspace{1.1pt}}|@{\hspace{1.1pt}}c@{\hspace{2.2pt}}c@{\hspace{1.1pt}}|@{\hspace{1.1pt}}c@{\hspace{2.2pt}}c@{\hspace{1.1pt}}|@{\hspace{1.1pt}}c@{\hspace{2.2pt}}c@{\hspace{2.2pt}}c@{\hspace{1.1pt}}|@{\hspace{1.1pt}}c@{\hspace{2.2pt}}c@{\hspace{2.2pt}}c@{}}
\toprule
& \multicolumn{4}{c|}{\textbf{Efficiency}} & \multicolumn{6}{c}{\textbf{Visual Quality}} \\
\cmidrule(lr){2-5}\cmidrule(l){6-11}
\textbf{Method} & \textbf{Latency (s)$\downarrow$} & \textbf{Speedup$\uparrow$} & \textbf{FLOPs (T)$\downarrow$} & \textbf{Speedup$\uparrow$} & \textbf{CLIPScore$\uparrow$} & \textbf{PickScore$\uparrow$} & \textbf{ImageReward$\uparrow$} & \textbf{PSNR$\uparrow$} & \textbf{SSIM$\uparrow$} & \textbf{LPIPS$\downarrow$} \\
\midrule
\textsc{Vanilla} ($N{=}50$) & 23.93 & 1.00 & 3734.6 & 1.00 & 32.54 & 23.77 & 1.361 & $\infty$ & 1.0000 & 0.0000 \\
\textsc{Vanilla} ($N{=}25$) & 11.58 & 2.06 & 1859.75 & 2.00 & 31.64 & 22.67 & 0.947 & 18.85 & 0.6818 & 0.2970 \\
\midrule
\textsc{$\Delta$-DiT} (i4) & 11.02 & 2.17 & 1602.9 & 2.33 & 32.65 & 23.38 & 1.241 & 14.60 & 0.6323 & 0.4909 \\
\textsc{ScalingCache} (FT6) & 10.93 & 2.19 & 1698.7 & 2.19 & 32.29 & 23.39 & 1.204 & 11.14 & 0.4934 & 0.5984 \\
\textsc{HiCache} (i4) & 10.59 & 2.26 & 1057.5 & 3.53 & 32.58 & \underline{23.80} & \underline{1.362} & \underline{19.52} & \underline{0.7320} & \underline{0.3031} \\
\textsc{TeaCache} ($\tau{=}.20$) & 8.63 & 2.77 & 1503.0 & 2.48 & \underline{32.73} & 23.73 & 1.349 & 16.89 & 0.7069 & 0.3675 \\
\textsc{FORA} (i4) & 8.12 & 2.95 & 983.2 & 3.80 & \textbf{32.79} & 23.55 & 1.294 & 15.26 & 0.6483 & 0.4516 \\
\textbf{\textsc{FeatFix}} ($N{=}50$) & 7.57 & 3.16 & 928.4 & 4.02 & 32.64 & \textbf{23.89} & \textbf{1.396} & \textbf{20.39} & \textbf{0.7534} & \textbf{0.2685} \\
\midrule
\textsc{TeaCache} ($\tau{=}.40$) & 5.56 & 4.30 & 759.3 & 4.92 & 32.75 & 23.66 & \underline{1.336} & \underline{16.49} & \textbf{0.6858} & \underline{0.4071} \\
\textsc{HiCache} (i8) & 5.35 & 4.47 & 611.4 & 6.11 & \underline{32.79} & 23.56 & 1.313 & 15.72 & 0.6266 & 0.4427 \\
\textsc{FORA} (i10) & 4.94 & 4.84 & 388.3 & 9.62 & 32.47 & 22.43 & 0.749 & 13.62 & 0.5415 & 0.6137 \\
\textsc{SpeCa} ($N{=}28$) & 4.80 & 4.99 & 618.7 & 6.04 & \textbf{32.95} & \underline{23.72} & 1.335 & 15.72 & 0.6367 & 0.4636 \\
\textsc{TaylorSeer} ($N{=}28$) & 4.77 & 5.02 & 536.5 & 6.96 & 32.55 & 23.37 & 1.273 & 15.55 & 0.6453 & 0.4615 \\
\textbf{\textsc{FeatFix}} ($N{=}28$) & 4.76 & 5.03 & 618.7 & 6.04 & 32.70 & \textbf{23.83} & \textbf{1.374} & \textbf{16.70} & \underline{0.6739} & \textbf{0.3953} \\
\midrule
\textsc{TeaCache} ($\tau{=}.55$) & 4.71 & 5.08 & 685.0 & 5.45 & 32.72 & \textbf{23.58} & 1.289 & 14.99 & 0.6063 & \underline{0.5138} \\
\textsc{HiCache} (i12) & 4.45 & 5.38 & 462.6 & 8.07 & 32.51 & 22.81 & 0.986 & 13.75 & 0.5379 & 0.5827 \\
\textsc{SpeCa} ($N{=}20$) & 3.73 & 6.41 & 467.2 & 7.99 & \textbf{33.08} & 23.54 & \underline{1.313} & \underline{15.03} & 0.6052 & 0.5163 \\
\textsc{TaylorSeer} ($N{=}20$) & 3.65 & 6.56 & 387.6 & 9.64 & 32.55 & 22.55 & 1.169 & 14.37 & \underline{0.6075} & 0.5189 \\
\textbf{\textsc{FeatFix}} ($N{=}20$) & 3.57 & 6.70 & 467.2 & 7.99 & \underline{33.03} & \underline{23.56} & \textbf{1.318} & \textbf{16.13} & \textbf{0.6166} & \textbf{0.4492} \\
\bottomrule
\end{tabular}
\caption{FLUX.1-dev on DrawBench200. Rules delimit speed regions; bold and underline mark the best and second-best quality within each region.}
\label{tab:flux_main}
\end{table*}

\subsection{Motivation for Whole-Block Correction}

Paid verification already produces the complete exact block-output tensor $[B,N,C]$. Forwarding only selected tokens or channels therefore saves no exact-block FLOPs and leaves an uncorrected residual. Attention heads are internal to the block and therefore do not provide a separate output axis for partial forwarding. At the other extreme, recomputing the whole timestep repeats the remaining transformer stack and changes the incoming trajectory. Whole-block forwarding thus reuses the complete paid output without this additional computation.

Figure~\ref{fig:motivation-correction}(a) reads from top to bottom. The upper diagrams compare token-, channel-, and whole-block forwarding. The middle annotations show that each block-level arm costs 467 TFLOPs, whereas whole-timestep recomputation costs 759 TFLOPs, or $1.63\times$ as much. The lower forest plot reports the additional downstream relative-$L_1$ reduction of whole-block forwarding over each partial arm, with all four 95\% confidence intervals lying above zero. Panel (b) tracks the Verify-minus-Whole-block difference in relative-$L_1$ across 200 paired DrawBench-style prompts, four layers, and early/middle/late interventions. Positive values indicate lower downstream error after correction. All 24 pointwise bootstrap intervals exclude zero, with an overall reduction of $2.541\,[2.254,2.835]\times10^{-3}$. Together, these results support whole-block forwarding as an effective local correction.

\subsection{FeatFix Framework}

\noindent\textbf{Overview.} Figure~\ref{fig:method_overview} situates the correction primitive within cached denoising. The top path follows state--block--feature flow from $x_T$ to $x_0$, where $T$ is the initial noise index and $L$ is the number of transformer blocks. Let $\mathcal S\subseteq\{1,\ldots,T\}\times\{1,\ldots,L\}$ denote the paid layer--timestep sites fixed before evaluation by a development-stage audit and the base caching schedule. Within each backbone and speed setting, the same sites are used for every evaluation sample and for both Verify and FeatFix. At a site $(t,\ell)\in\mathcal S$, Verify and FeatFix evaluate the same incoming state $u_{t,\ell}$, draft $h^{\mathrm d}_{t,\ell}$, and exact feature $h^{\mathrm f}_{t,\ell}$. For compactness, we omit site subscripts and write $u$, $h_{\rm d}$, and $h_{\rm f}$. In the primary feature analyses, the figure's $d(h_{\rm d},h_{\rm f})$ denotes Eq.~\eqref{eq:relative-l1}. Metric ablations may replace this logged scalar without changing the fixed forwarding rule. We write $h^{\mathrm{out}}_{t,\ell}$ for the block output forwarded downstream. Verify sets $h^{\mathrm{out}}_{t,\ell}=h^{\mathrm d}_{t,\ell}$ and discards the paid exact tensor, whereas FeatFix sets $h^{\mathrm{out}}_{t,\ell}=h^{\mathrm f}_{t,\ell}$. The shorthand $h^{\mathrm{out}}=h_{\rm f}$ is local to the corrected site under the same incoming state. Outside the selected verification sites, FeatFix preserves the underlying draft policy.

\noindent\textbf{Local feature correction.} At each $(t,\ell)\in\mathcal S$, FeatFix logs the discrepancy for analysis and applies the fixed forwarding rule below. The schedule is fixed before evaluation and shared across evaluation samples within each setting.
\begin{equation}
h^{\mathrm{out}}_{t,\ell}=\begin{cases}
h^{\mathrm f}_{t,\ell}, &(t,\ell)\in\mathcal S \quad\text{(FeatFix)},\\
\mathcal D(u_{t,\ell}), &(t,\ell)\notin\mathcal S.
\end{cases}
\label{eq:correction}
\end{equation}
Thus replacement eliminates the local same-input draft residual by construction. The method forwards the complete block-output feature tensor, typically $[B,N,C]$, rather than a subset of tokens, channels, or attention heads. Because the paid exact branch has already produced the whole tensor, whole-block forwarding adds no exact-block FLOPs relative to Verify. Partial replacement leaves an uncorrected residual, while whole-timestep recomputation repeats substantially more transformer work. Whole-block forwarding is therefore the compute-consistent sparse intervention. The mechanism analysis below tests its local reset and downstream propagation.

\noindent\textbf{Matched-compute controls and accounting.} Every run separately records full-refresh timesteps, paid exact-block checks, and forwarded corrections. Verify uses the same configured full-refresh and paid-check budgets and forwards $h^{\mathrm d}$ at each paid site. At the strict FLUX settings up to $6.70\times$, realized sites match for every prompt, providing a controlled same-site comparison. When every paid exact output is forwarded, Exact-only removes the redundant draft and discrepancy paths and produces outputs identical to FeatFix in the implementation audit. This confirms implementation parity between the two execution paths.

\section{Experiments}

We first compare FeatFix with cache and forecasting baselines on FLUX, then isolate the effect of forwarding the verified feature under matched compute. Mechanism studies measure the local reset and downstream error propagation, while HunyuanVideo-1.5, Qwen-Image, and DiT-XL/2 test portability.

\subsection{Experimental Setup}

FLUX.1-dev~\cite{blackforest2024flux} is the primary backbone. The main evaluation uses a fixed $200$-prompt DrawBench list~\cite{saharia2022imagen} with matched order and seeds, while the powered study uses $900$ normalized COCO-caption clusters~\cite{lin2014coco} with two seeds each. We report efficiency (latency, speedup, and FLOPs), reconstruction (PSNR, SSIM, and LPIPS), and text-alignment and preference metrics (CLIPScore, PickScore, and ImageReward)~\cite{wang2004ssim,zhang2018lpips,hessel2021clipscore,kirstain2023pickscore,xu2023imagereward}. DiT-XL/2 on ImageNet-256, Qwen-Image, and HunyuanVideo-1.5 test cross-model transfer, with FID, IS, precision, recall, GenEval, and VBench reported where applicable~\cite{peebles2023dit,deng2009imagenet,qwenimage2025,hunyuanvideo2025,heusel2017fid,ghosh2023geneval,huang2024vbench}. Paired FLUX intervals use prompt- or cluster-level bootstrap resampling.

\paragraph{Evaluation protocol.} We compare training-free methods within their recorded timing cohorts, with Vanilla as an unranked reference; within near-speed groups, bold and underline mark the best and second-best quality results. The primary causal comparison is paired FeatFix versus Verify with matched full-refresh and paid-site traces.

\begin{figure}[!t]
\centering
\includegraphics[width=\columnwidth]{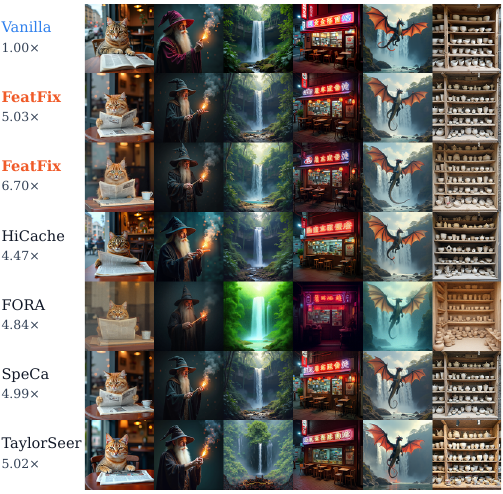}
\caption{FLUX.1-dev visual comparison.}
\label{fig:flux-baseline-visual}
\end{figure}

\begin{table}[!t]
\centering
\small
\begin{tabular}{@{}l@{\hspace{0.7pt}}|@{\hspace{0.7pt}}c@{\hspace{0.7pt}}|@{\hspace{0.7pt}}c@{\hspace{1.4pt}}c@{\hspace{1.4pt}}c@{\hspace{0.7pt}}|@{\hspace{0.7pt}}c@{\hspace{1.4pt}}c@{\hspace{1.4pt}}c@{}}
\toprule
\textbf{Arm} & \textbf{Lat. (s)$\downarrow$} & \textbf{CLIP$\uparrow$} & \textbf{Pick$\uparrow$} & \textbf{IR$\uparrow$} & \textbf{PSNR$\uparrow$} & \textbf{SSIM$\uparrow$} & \textbf{LPIPS$\downarrow$} \\
\midrule
\textsc{Vanilla} & 23.93 & 32.54 & 23.77 & 1.361 & $\infty$ & 1.0000 & 0.0000 \\
\midrule
\textsc{Verify} & 3.60 & 31.48 & 22.80 & 0.911 & 14.29 & 0.5854 & 0.5381 \\
\textbf{\textsc{FeatFix}} & 3.57 & \textbf{33.03} & \textbf{23.56} & \textbf{1.318} & \textbf{16.13} & \textbf{0.6166} & \textbf{0.4492} \\
\bottomrule
\end{tabular}
\caption{FLUX fixed-fallback comparison between Vanilla, Verify, and FeatFix.}
\label{tab:flux_powered_main}
\end{table}

\subsection{Main Results on FLUX}

\noindent\textbf{Overall FLUX comparison.} Table~\ref{tab:flux_main} compares FeatFix with reuse, caching, and forecasting baselines in three speed regions. The horizontal rules group methods with similar latency, which makes the quality comparisons within each region more meaningful. FeatFix is among the fastest methods in every group and reaches a $6.70\times$ speedup in the most aggressive setting. This range shows that FeatFix can be used for both moderate and aggressive acceleration.

Among the accelerated methods, FeatFix gives the best ImageReward in all three regions and leads most PickScore and reconstruction comparisons. Its advantage is especially clear when acceleration becomes more aggressive, where several baselines lose both perceptual quality and similarity to Vanilla. A few methods retain a small advantage on CLIPScore or SSIM in individual regions, so FeatFix is not the best accelerated entry for every metric. However, its results are more balanced across preference and reconstruction measures.

\noindent\textbf{FLUX qualitative comparison.} Figure~\ref{fig:flux-baseline-visual} shows the corresponding FLUX examples. At both displayed speed settings, FeatFix keeps the main objects, composition, colors, and local details close to Vanilla. The examples support the quantitative results by showing that the quality gains are visible across different prompts rather than being limited to one image type.

\begin{figure}[!t]
\centering
\begin{tabular}{@{}p{0.485\columnwidth}@{\hspace{0.03\columnwidth}}p{0.485\columnwidth}@{}}
\centering\includegraphics[width=\linewidth]{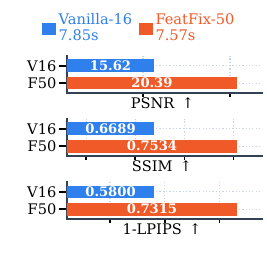} & \centering\includegraphics[width=\linewidth]{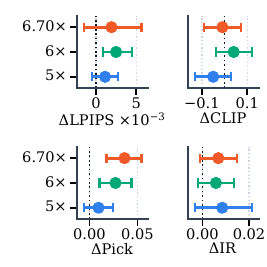}\tabularnewline[2pt]
\centering{\small\textbf{(a) Runtime-matched reference.}} & \centering{\small\textbf{(b) Metric-wise paired effects.}}\tabularnewline
\end{tabular}
\caption{FLUX runtime-matched reference and paired effects across three regions up to $6.70\times$.}
\label{fig:regime-flux}
\end{figure}

\begin{figure}[!t]
\centering
\begin{tabular}{@{}p{0.485\columnwidth}@{\hspace{0.03\columnwidth}}p{0.485\columnwidth}@{}}
\centering\includegraphics[width=\linewidth]{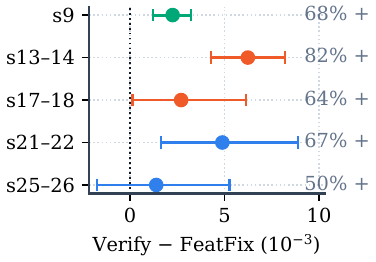} & \centering\includegraphics[width=\linewidth]{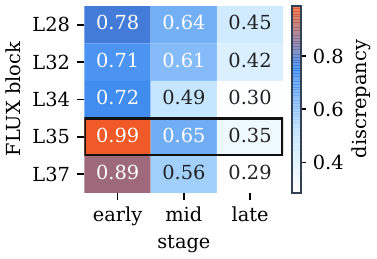}\tabularnewline[2pt]
\centering{\small\textbf{(a) Stagewise propagation.}} & \centering{\small\textbf{(b) Layer--stage drift map.}}\tabularnewline
\end{tabular}
\caption{FLUX correction propagation across denoising stages and measured layers.}
\label{fig:mechanism-propagation}
\end{figure}

\begin{figure}[!t]
\centering
\begin{tabular}{@{}p{0.485\columnwidth}@{\hspace{0.03\columnwidth}}p{0.485\columnwidth}@{}}
\centering\includegraphics[width=\linewidth]{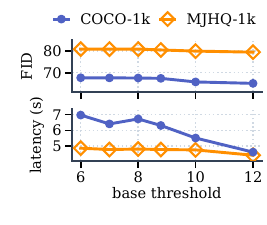} & \centering\includegraphics[width=\linewidth]{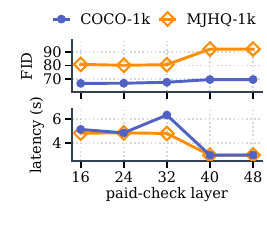}\tabularnewline[2pt]
\centering{\small\textbf{(a) Threshold sensitivity.}} & \centering{\small\textbf{(b) Paid-layer sensitivity.}}\tabularnewline
\end{tabular}
\caption{FLUX threshold and paid-layer sensitivity.}
\label{fig:regime-flux-sensitivity}
\end{figure}

\begin{figure}[!t]
\centering
\begin{tabular}{@{}p{0.485\columnwidth}@{\hspace{0.03\columnwidth}}p{0.485\columnwidth}@{}}
\centering\includegraphics[width=\linewidth]{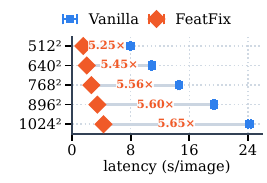} & \centering\includegraphics[width=\linewidth]{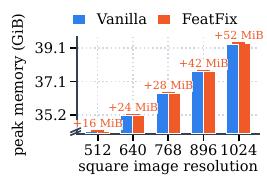}\tabularnewline[2pt]
\centering{\small\textbf{(a) Five-resolution latency.}} & \centering{\small\textbf{(b) Peak-memory scaling.}}\tabularnewline
\end{tabular}
\caption{FLUX latency and peak-memory scaling across five resolutions.}
\label{fig:flux-resolution-scaling}
\end{figure}

\begin{figure}[!t]
\centering
\includegraphics{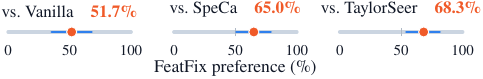}
\caption{FLUX.1-dev 2AFC preference with 95\% confidence intervals and a 50\% baseline.}
\label{fig:human-preference}
\end{figure}

\noindent\textbf{Fixed-fallback comparison.} Table~\ref{tab:flux_powered_main} directly compares FeatFix with the matched Verify control. The two methods have nearly identical latency and use the same paid exact checks. FeatFix performs better on every reported quality metric because it forwards the exact feature instead of discarding it. This comparison isolates the effect of correction from the underlying draft schedule.

\noindent\textbf{Matched-runtime and paired effects.} Figure~\ref{fig:regime-flux}(a) places a lower-step Vanilla reference next to FeatFix at a similar observed runtime. FeatFix gives better reconstruction in this runtime range. Figure~\ref{fig:regime-flux}(b) provides the paired comparison with Verify. The effects remain favorable across all three speed regions: FeatFix improves ImageReward and LPIPS while changing latency only slightly.

\subsection{Mechanism Analysis}
\label{sec:correction-value}

\noindent\textbf{Downstream feature propagation.} Figure~\ref{fig:mechanism-propagation} tracks the effect of correction through later denoising stages and layers. Panel (a) shows that the reduction in feature error remains positive through most of the later stages instead of disappearing immediately after the corrected site. Panel (b) reports the measured layers directly and shows the same pattern around the correction anchor.

\noindent\textbf{Threshold and layer sensitivity.} Figure~\ref{fig:regime-flux-sensitivity}(a) varies the base threshold on two evaluation sets. Increasing the threshold reduces latency, but quality does not improve monotonically, so the fastest configuration is not always the best operating point. Panel (b) changes the paid layer and shows a similar tradeoff. Some layers provide a better balance between speed and quality, which supports calibrating the correction site.

\noindent\textbf{Resolution scaling.} Figure~\ref{fig:flux-resolution-scaling} measures latency and peak memory from $512^2$ to $1024^2$. The latency gap between Vanilla and FeatFix grows with resolution, while the resulting speedup stays in a narrow range. Peak memory also grows for both methods, but the additional memory used by FeatFix remains small. These results show that the correction mechanism scales without introducing a large memory cost.

\noindent\textbf{Human Preference Study.} Figure~\ref{fig:human-preference} summarizes a blinded two-alternative forced-choice study in which 18 participants each answered 10 questions. FeatFix is preferred to the near-speed SpeCa and TaylorSeer settings, while its comparison with Vanilla stays close to the 50\% chance line. The confidence intervals in the figure show the uncertainty of each comparison. Together, the human study provides further evidence that FeatFix preserves perceptual quality under acceleration and improves over near-speed baselines.

\begin{table}[!t]
\centering
\small
\begin{tabular}{@{}l@{\hspace{1pt}}|@{\hspace{1pt}}c@{\hspace{2pt}}c@{\hspace{1pt}}|@{\hspace{1pt}}c@{\hspace{1pt}}|@{\hspace{1pt}}c@{\hspace{2pt}}c@{\hspace{2pt}}c@{}}
\toprule
& \multicolumn{2}{c|}{\textbf{Efficiency}} & \multicolumn{4}{c}{\textbf{Visual Quality}} \\
\cmidrule(lr){2-3}\cmidrule(l){4-7}
\textbf{Method} & \textbf{Lat. (s)$\downarrow$} & \textbf{Spd.$\uparrow$} & \textbf{VBench$\uparrow$} & \textbf{PSNR$\uparrow$} & \textbf{SSIM$\uparrow$} & \textbf{LPIPS$\downarrow$} \\
\midrule
\textsc{Vanilla} & 772.60 & 1.00 & 81.19 & $\infty$ & 1.0000 & 0.0000 \\
\midrule
\textsc{FasterCache} & 510.10 & 1.51 & 81.57 & 23.34 & 0.8419 & 0.1459 \\
\textsc{ToCa} & 491.20 & 1.57 & 81.51 & \underline{24.80} & \underline{0.8569} & \underline{0.1316} \\
\textsc{DiCache} & 487.70 & 1.58 & \underline{81.65} & 22.51 & 0.7983 & 0.1939 \\
\textbf{\textsc{FeatFix}} & 483.50 & 1.60 & \textbf{82.09} & \textbf{26.53} & \textbf{0.8849} & \textbf{0.1035} \\
\midrule
\textsc{SpeCa} & 444.40 & 1.74 & 81.41 & 23.36 & 0.8231 & 0.1663 \\
\textsc{DBCache} & 435.30 & 1.78 & \underline{81.65} & 24.62 & 0.8547 & 0.1343 \\
\textsc{MagCache} & 418.30 & 1.85 & 81.50 & \underline{25.03} & \underline{0.8643} & \underline{0.1237} \\
\textsc{PAB} & 416.80 & 1.85 & \underline{81.65} & 23.69 & 0.8185 & 0.1689 \\
\textbf{\textsc{FeatFix}} & 415.10 & 1.86 & \textbf{82.15} & \textbf{26.06} & \textbf{0.8739} & \textbf{0.1154} \\
\midrule
\textsc{DeepCache} & 358.20 & 2.16 & \underline{81.45} & \underline{23.56} & \underline{0.8335} & \underline{0.1574} \\
\textsc{TeaCache} & 356.80 & 2.17 & 81.30 & 23.14 & 0.8242 & 0.1672 \\
\textsc{TaylorSeer} & 342.10 & 2.26 & 80.18 & 14.76 & 0.6045 & 0.4154 \\
\textbf{\textsc{FeatFix}} & 338.40 & 2.28 & \textbf{82.03} & \textbf{23.86} & \textbf{0.8385} & \textbf{0.1511} \\
\bottomrule
\end{tabular}
\caption{HunyuanVideo-1.5 65-frame transfer across three near-speed regions with scaled VBench and reconstruction metrics.}
\label{tab:hunyuan_transfer}
\end{table}

\subsection{Cross-Backbone Generalization}

\noindent\textbf{HunyuanVideo-1.5 results.} Table~\ref{tab:hunyuan_transfer} compares FeatFix with video acceleration baselines in three speed regions. Within each region, the methods have similar latency, so the main difference is the quality of the generated videos. Among the accelerated methods, FeatFix gives the best VBench and reconstruction results in all three groups. The advantage remains visible at the fastest setting, where some baselines show a larger drop in reconstruction quality.

\begin{figure}[!t]
\centering
\includegraphics[width=\linewidth]{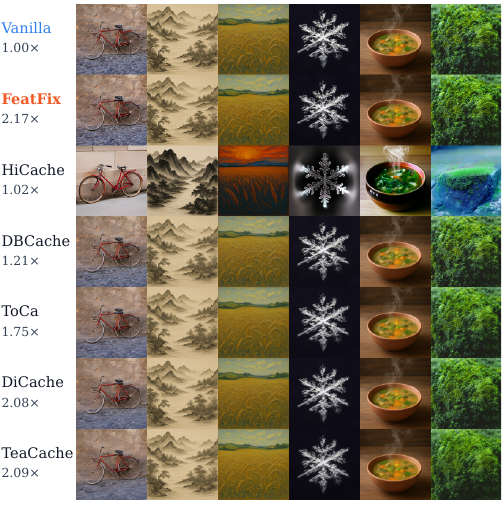}
\caption{Qwen-Image visual comparison.}
\label{fig:qwen-baseline-visual}
\end{figure}

\noindent\textbf{Qwen-Image results.} Figure~\ref{fig:qwen-baseline-visual} compares Qwen-Image outputs at two speed settings. FeatFix retains the main subjects, layouts, and textures of the Vanilla images while reducing generation time, whereas several cache baselines change object appearance or local details more strongly. Table~\ref{tab:qwen_transfer} reports the corresponding results in two speed regions. Among the accelerated methods, FeatFix gives the strongest PickScore and ImageReward results in both groups and also reaches the top CLIPScore. DBCache and DiCache obtain better reconstruction scores at some operating points, indicating closer pixel-level agreement with the corresponding Vanilla images. Together, the visual and quantitative results show that FeatFix provides a stronger balance on preference and prompt-alignment metrics.

\begin{figure}[!t]
\centering
\includegraphics[width=\linewidth]{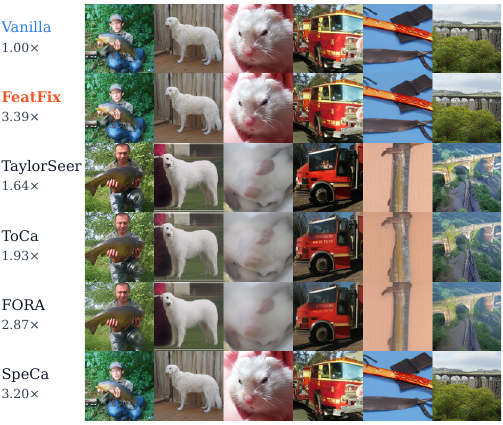}
\caption{DiT-XL/2 visual comparison.}
\label{fig:dit-baseline-visual}
\end{figure}

\noindent\textbf{DiT-XL/2 results.} Figure~\ref{fig:dit-baseline-visual} compares class-conditional samples from DiT-XL/2. FeatFix preserves the main object identity, shape, and texture of the Vanilla samples, while several faster baselines introduce larger changes in object structure or background content. Table~\ref{tab:dit_transfer} compares the same methods on ImageNet-256 using distributional quality and precision--recall metrics. FeatFix is the fastest entry and, among the accelerated methods, gives the best FID, sFID, IS, and precision. Together, these results show that the correction remains effective on a class-conditional diffusion transformer and is not limited to text-conditioned image or video models.

\begin{table}[!t]
\centering
\small
\begin{tabular}{@{}l@{\hspace{0.7pt}}|@{\hspace{0.7pt}}c@{\hspace{0.7pt}}|@{\hspace{0.7pt}}c@{\hspace{1.4pt}}c@{\hspace{1.4pt}}c@{\hspace{0.7pt}}|@{\hspace{0.7pt}}c@{\hspace{1.4pt}}c@{\hspace{1.4pt}}c@{}}
\toprule
& \textbf{Efficiency} & \multicolumn{6}{c}{\textbf{Visual Quality}} \\
\cmidrule(lr){2-2}\cmidrule(l){3-8}
\textbf{Method} & \textbf{Lat. (s)$\downarrow$} & \textbf{CLIP$\uparrow$} & \textbf{Pick$\uparrow$} & \textbf{IR$\uparrow$} & \textbf{PSNR$\uparrow$} & \textbf{SSIM$\uparrow$} & \textbf{LPIPS$\downarrow$} \\
\midrule
\textsc{Vanilla} & 22.59 & 32.56 & 23.01 & 1.174 & $\infty$ & 1.0000 & 0.0000 \\
\midrule
\textsc{HiCache} & 22.15 & 32.38 & 22.65 & 1.084 & 10.19 & 0.3848 & 0.7573 \\
\textsc{DBCache} & 18.74 & 32.60 & \underline{22.99} & 1.158 & \textbf{38.62} & \underline{0.9802} & \textbf{0.0171} \\
\textsc{TeaCache} & 15.56 & \textbf{32.71} & 22.90 & 1.129 & 29.43 & 0.9355 & 0.0896 \\
\textsc{DiCache} & 15.42 & \underline{32.62} & \underline{22.99} & \underline{1.168} & \underline{38.03} & \textbf{0.9809} & \underline{0.0186} \\
\textbf{\textsc{FeatFix}} & 15.08 & \textbf{32.71} & \textbf{23.03} & \textbf{1.194} & 27.41 & 0.9169 & 0.1116 \\
\midrule

\textsc{ToCa} & 12.90 & 32.57 & 22.87 & \underline{1.106} & \textbf{32.46} & \underline{0.9460} & \underline{0.0754} \\
\textsc{DiCache} & 10.84 & \underline{32.63} & \underline{22.92} & 1.105 & \underline{32.32} & \textbf{0.9565} & \textbf{0.0593} \\
\textsc{TeaCache} & 10.83 & 32.60 & 22.72 & 1.012 & 25.80 & 0.8973 & 0.1601 \\
\textbf{\textsc{FeatFix}} & 10.40 & \textbf{32.76} & \textbf{22.94} & \textbf{1.154} & 26.60 & 0.9069 & 0.1274 \\
\bottomrule
\end{tabular}
\caption{Qwen-Image transfer on DrawBench200 across two speed regions; rules delimit within-region comparisons.}
\label{tab:qwen_transfer}
\end{table}

\begin{table}[!t]
\centering
\small
\begin{tabular}{@{}l@{\hspace{1pt}}|@{\hspace{1pt}}c@{\hspace{1pt}}|@{\hspace{1pt}}c@{\hspace{2pt}}c@{\hspace{2pt}}c@{\hspace{1pt}}|@{\hspace{1pt}}c@{\hspace{2pt}}c@{}}
\toprule
& \multicolumn{1}{c|}{\textbf{Efficiency}} & \multicolumn{5}{c}{\textbf{Visual Quality}} \\
\cmidrule(lr){2-2}\cmidrule(l){3-7}
\textbf{Method} & \textbf{Lat. (s)$\downarrow$} & \textbf{FID$\downarrow$} & \textbf{sFID$\downarrow$} & \textbf{IS$\uparrow$} & \textbf{Precision$\uparrow$} & \textbf{Recall$\uparrow$} \\
\midrule
\textsc{Vanilla} & 1.52 & 2.585 & 33.90 & 243.3 & 0.8058 & 0.5667 \\
\midrule
\textsc{TaylorSeer} & 0.93 & 3.016 & -- & 235.1 & 0.8068 & 0.5574 \\
\textsc{ToCa} & 0.79 & 3.804 & -- & 226.5 & 0.7979 & 0.5336 \\
\textsc{FORA} & 0.53 & 5.132 & -- & 212.5 & 0.7870 & 0.5024 \\
\textsc{SpeCa} & 0.48 & \underline{2.807} & \underline{30.94} & \underline{238.8} & \underline{0.8084} & \textbf{0.5595} \\
\textbf{\textsc{FeatFix}} & 0.45 & \textbf{2.792} & \textbf{30.51} & \textbf{240.4} & \textbf{0.8115} & \underline{0.5577} \\
\bottomrule
\end{tabular}
\caption{DiT-XL/2 transfer on ImageNet-256 with distributional and precision--recall metrics.}
\label{tab:dit_transfer}
\end{table}

\section{Limitations}

FeatFix focuses on a common setting in which cache- or forecast-based accelerators already evaluate occasional exact block features. This focus enables a clean matched-compute study while keeping the correction mechanism training-free, local, and auditable. For the reported settings, we freeze the sites after development-stage audits and keep them fixed within each evaluation. A natural next step is to extend this principle to automatic budget-aware site selection and additional inference stacks.

\section{Conclusion}

We presented FeatFix, a local exact-feature correction method for cached diffusion inference. At frozen layer--timestep sites, it reused already-computed exact block features, replaced selected draft features, preserved the underlying draft policy, and avoided full-timestep recomputation. Matched Verify controls and feature traces showed that this intervention reset the same-input residual and reduced downstream error at essentially unchanged latency. Across FLUX.1-dev, Qwen-Image, HunyuanVideo-1.5, and DiT-XL/2, FeatFix reached up to $6.70\times$ speedup over Vanilla and maintained competitive output quality. In the controlled FLUX study, FeatFix improved ImageReward and passed the LPIPS non-inferiority guard under matched computation. Together, these results established that paid verification served as both an audit signal and a practical resource for local trajectory correction. Building on this foundation, future work will explore prompt-adaptive, budget-aware policies that dynamically allocate exact-feature corrections across layers and timesteps.

\bibliography{references}

\end{document}